# Reusing Convolutional Activations from Frame to Frame to Speed up Training and Inference

**Arno Khachatourian**
arno.kha@gmail.com

## Abstract

When processing similar frames in succession, we can take advantage of the locality of the convolution operation to reevaluate only portions of the image that changed from the previous frame. By saving the output of a layer of convolutions and calculating the change from frame to frame, we can reuse previous activations and save computational resources that would otherwise be wasted recalculating convolutions whose outputs we have already observed. This technique can be applied to many domains, such as processing videos from stationary video cameras, studying the effects of occluding or distorting sections of images, applying convolution to multiple frames of audio or time series data, or playing Atari games. Furthermore, this technique can be applied to speed up both training and inference.

## 1 Discussion

Convolutional layers serve as an essential building block for most modern image processing neural networks [1], serving as higher-level feature extractors [2]. We can take advantage of the locality and separability of convolutions to perform the operations in parallel [3], greatly speeding up the computation of the neural network. However, we can also take advantage of the locality of convolutions to only reprocess changed areas in the input layer, as they only affect a small local area in the next layer. To accomplish this, we simply store the convolutional activations of the previous frame, and reuse the activations in the successive frame wherever the corresponding input pixels have not changed. This is possible in frameworks such as PyTorch, which use dynamic computational graphs, allowing one to define the computational graph on each iteration of training or inference. This technique can be employed to save computation in any application where there are large areas of input that remain the same from frame to frame, such as processing video from stationary cameras, training to recognize a pattern instance at different locations, training autoencoders to restore images after inpainting [4], and playing Atari games. For certain applications such as studying the effects of partial occlusion, neural inpainting, or training the network to recognize a pattern at multiple locations, we do not even have to calculate the portions of the image that have changed, since we are defining those ourselves. Furthermore, using the last $n$ frames as input will be less expensive with this technique, since the outputs for the previous $n-1$ frames have already been calculated.

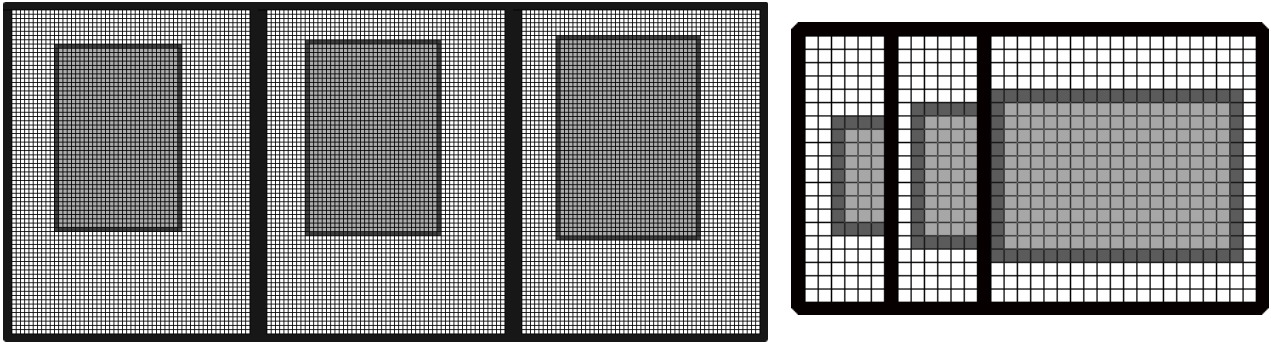

**Figure 1**: The effects of a changed area of pixels in an input image on the next two layers of convolution. Here, a convolution with a kernel size of three is shown on (a) an 80x60 image, and (b) a 20x20 image. In general, convolutions of kernel size $k$ only affect the $(k-1)/2$ pixels on either side of an input pixel in the next layer.

Convolutional neural network architectures are continuing to grow and are taking longer to train [5], increasing the necessity of GPUs for neural network training and inference. Using a CPU and reusing convolutional activations can be a viable alternative in some scenarios. This method can allow individuals without GPUs to still train or run neural networks without suffering too much of a cost in terms of speed. Additionally, once deployed, this technique can allow models to perform inference more cheaply, since they do not require GPUs. Furthermore, depending on the specific architecture and use case, these methods can be faster, which could impact real-time systems. Reusing activations on a CPU also allows for the use of RAM instead of dedicated VRAM on a GPU, allowing for more model parameters or larger inputs to the network due to a larger, cheaper memory pool.

One of the disadvantages that this method currently faces is that it is slower to run on a GPU than a regular convolutional network. Since GPU implementations are highly parallelized, processing more pixels per frame comes at a marginal cost. In addition, there are many cases where the overhead required to perform these reuse operations exceed the benefit they provide, namely when there is not enough similarity in input at each successive step. Another limitation to this method is that as the number of model parameters grows, the worse convolutional reuse on a CPU performs in comparison to the baseline method on a GPU, although its performance in comparison to the baseline method on a CPU greatly improves. Even though the reuse method is mitigating more computation in these cases, there is also more remaining computation to execute.

## 2 Results

I timed the reuse model on a CPU against a base model on both a CPU and GPU in the OpenAI Gym Atari [6] environment. Both training and inference speed were evaluated on an architecture consisting of two convolutional layers followed by a dense layer. These layers comprised a policy that was trained using the REINFORCE algorithm. The input images were converted to grayscale and downsampled by a factor of either two or four in each dimension.

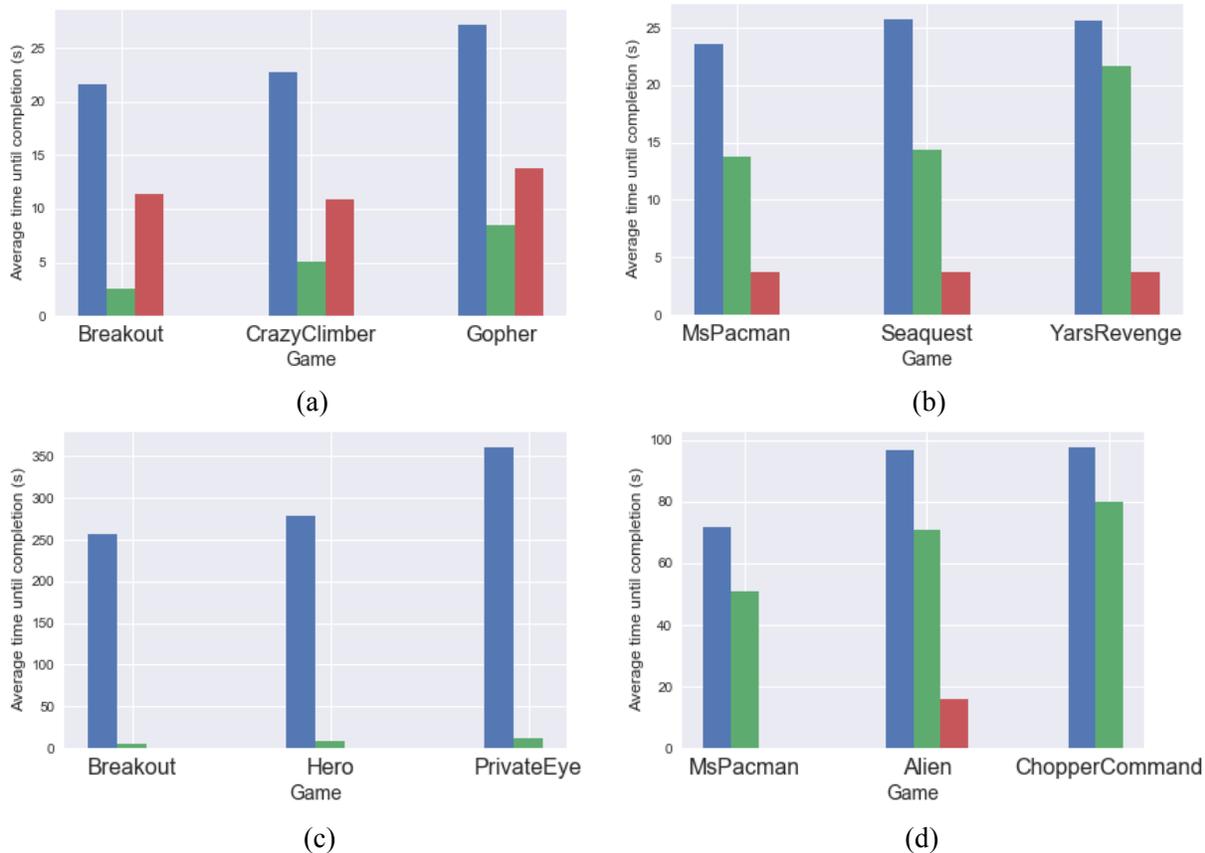

**Figure 2**: The best (left) and worst (right) of results for inference (top) and training (bottom) compared to the base model, using 40 filters for layer one and 80 filters for layer two. The CPU version of the base model is represented by the blue bars, the reuse model is represented by the green bars, and the GPU version of the base model is represented by the red bars. The missing red bars in (c) and (d) are due to the GPU running out of memory and the test terminating before completion.

For most test settings, the reuse model was much more effective than the base model run on the CPU. Compared to the base model on the GPU, the reuse model was about as fast for inference when the input image was downsampled by a factor of four. When the image was downsampled by a factor of two, the reuse model performed about as well when the number of parameters were small, then got slightly slower as the number of model parameters grew. For training, the reuse model was slower than the base GPU model, but was able to process larger images and use more model parameters. The results have high variance and are highly dependent on the level of similarity of the input from frame to frame. There are some games for which the reuse model was better in every test setting, such as Breakout, and some games for which the reuse model was worse in every test setting, such as BattleZone.

In summary, the reuse model worked well when there were small areas of pixel change from frame to frame, and performed poorly when those areas of change were large. As the number of the filters in each layer grows, the reuse model outscales the base version run on the CPU, but is outscaled by the base model run on the GPU. The complete result tables are available in the Appendix. The code used to run the experiments can be found here:

```
https://github.com/arnokha/reusing_convolutions/tree/master/performance_tests
```

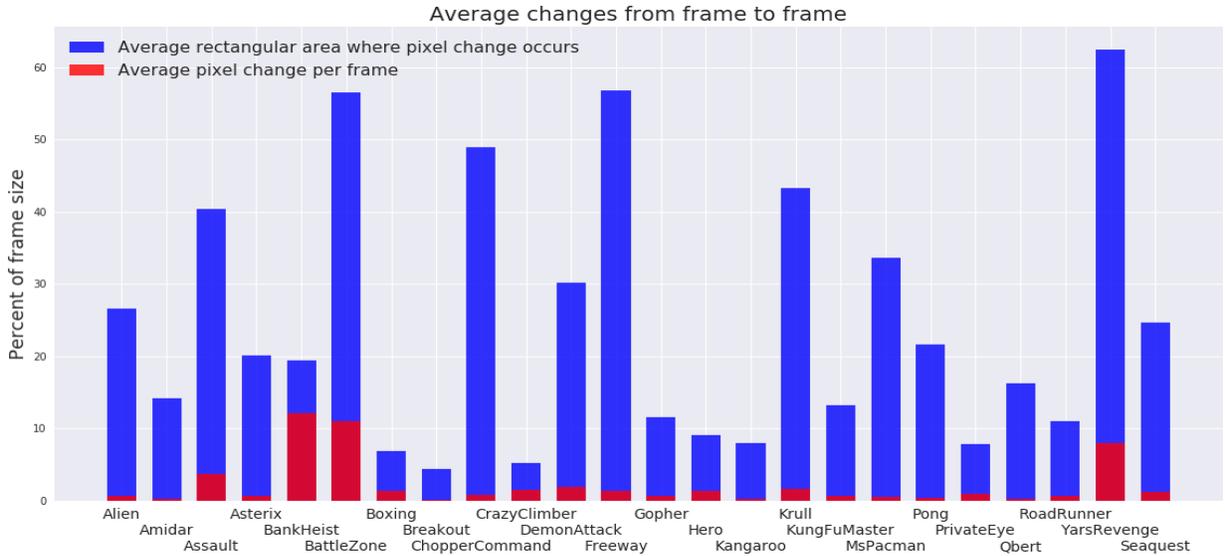

**Figure 3**: Average number of pixel changes from frame to frame in Atari games. These were obtained by averaging over 50,000 observations, taking random actions at each time step.

## 3 Future Work

There are many promising variations and applications of convolutional reuse to explore.
Using multiple input frames is a common approach in neural networks [7,8] and should provide even more computational savings. Audio and time series data are also candidates for this kind of application, alongside video.

Detecting changes in small local areas and recalculating only those areas may prove much more efficient than recalculating one large rectangular area every frame. Figure 3 shows that the rectangular area used for recalculation is much larger than the actual number of pixels changed in each frame. There may also be opportunity to detect pixel shifts using optical flow or other methods and match those shifts by shifting the convolutional activations.

It would also be instructive to measure the power usage of this method, especially on GPUs. Though it may be slightly slower than automatically recalculating convolutions at every step, there is a possibility that it may provide considerable power reduction.

## 4 Conclusion

Reusing convolutional activations with CPUs is a good way to save computation for both training and inference, and can serve as a viable alternative to training or doing inference on GPUs in some scenarios. It is likely cheaper, sometimes faster, and it will likely have access to more memory. Unfortunately, there is currently not as much incentive to use this method on GPUs, other than possibly saving power. There are many possible application domains for this technique, and there are likely many ways to improve upon it.

**Appendix: Tables of results and hardware specifications**

CPU: Intel Core i7-7700K Kaby Lake with 32GB RAM
GPU: NVIDIA 1080ti

The times that appear in the tables are averaged over 10 runs. All times are in seconds. The mean times over all games are listed in the bottom row of the table

The tables available in HTML form here:
`https://github.com/arnokha/reusing_convolutions/tree/master/figures`

Inference: time to run 3000 steps, grayscale, downsampled by a factor of 4 in each dimension

Number of filters in convolutional layers one and two: 20/40

| | base_cpu | reuse_cpu | base_gpu | reuse_gpu |
|---|---|---|---|---|
| Alien | 3.9 | 3.3 | 3.2 | 3.9 |
| Amidar | 3.7 | 2.2 | 3.1 | 3.5 |
| Assault | 3.2 | 3.0 | 2.8 | 3.6 |
| Asterix | 2.8 | 2.1 | 2.6 | 3.0 |
| BankHeist | 3.8 | 3.2 | 3.0 | 3.8 |
| BattleZone | 3.5 | 3.6 | 2.9 | 3.6 |
| Boxing | 3.7 | 2.8 | 3.1 | 3.7 |
| Breakout | 3.0 | 1.6 | 2.4 | 2.3 |
| ChopperCommand | 3.5 | 3.4 | 3.0 | 3.6 |
| CrazyClimber | 2.8 | 1.6 | 2.6 | 2.7 |
| DemonAttack | 2.8 | 2.5 | 2.3 | 3.2 |
| Freeway | 3.5 | 3.8 | 3.0 | 3.8 |
| Gopher | 3.1 | 2.1 | 2.9 | 3.4 |
| Hero | 3.6 | 2.0 | 3.0 | 2.9 |
| Kangaroo | 3.5 | 2.5 | 2.9 | 3.4 |
| Krull | 3.6 | 3.4 | 3.0 | 3.6 |
| KungFuMaster | 3.4 | 2.3 | 3.0 | 3.5 |
| MsPacman | 3.1 | 2.9 | 2.9 | 3.5 |
| Pong | 2.9 | 2.1 | 3.1 | 4.7 |
| PrivateEye | 3.5 | 2.3 | 3.0 | 3.4 |
| Qbert | 3.1 | 1.8 | 2.5 | 2.5 |
| RoadRunner | 3.5 | 2.2 | 3.0 | 3.3 |
| Seaquest | 3.2 | 2.6 | 2.8 | 3.3 |
| YarsRevenge | 3.1 | 3.4 | 2.7 | 3.4 |
| Mean | 3.3 | 2.6 | 2.9 | 3.4 |

Number of filters in convolutional layers one and two: 40/80

| | base_cpu | reuse_cpu | base_gpu | reuse_gpu |
|---|---|---|---|---|
| Alien | 7.1 | 5.4 | 3.9 | 5.0 |
| Amidar | 7.5 | 3.2 | 4.4 | 4.7 |
| Assault | 6.8 | 4.8 | 4.5 | 5.0 |
| Asterix | 5.9 | 4.0 | 3.6 | 4.1 |
| BankHeist | 7.8 | 5.0 | 4.4 | 4.9 |
| BattleZone | 6.8 | 5.8 | 3.8 | 4.6 |
| Boxing | 7.0 | 3.9 | 3.8 | 4.7 |
| Breakout | 5.7 | 1.9 | 3.8 | 3.9 |
| ChopperCommand | 6.8 | 4.8 | 3.8 | 4.6 |
| CrazyClimber | 5.8 | 2.0 | 3.6 | 3.8 |
| DemonAttack | 5.7 | 3.0 | 3.8 | 4.3 |
| Freeway | 6.1 | 6.2 | 4.0 | 4.8 |
| Gopher | 6.9 | 2.8 | 4.6 | 4.8 |
| Hero | 6.8 | 2.6 | 3.9 | 4.1 |
| Kangaroo | 6.7 | 3.5 | 3.8 | 4.2 |
| Krull | 6.7 | 5.2 | 3.8 | 4.5 |
| KungFuMaster | 6.5 | 3.6 | 4.0 | 4.5 |
| MsPacman | 6.1 | 4.6 | 3.7 | 4.5 |
| Pong | 5.8 | 2.7 | 4.4 | 5.6 |
| PrivateEye | 6.9 | 4.2 | 3.8 | 4.7 |
| Qbert | 6.0 | 1.9 | 3.9 | 4.0 |
| RoadRunner | 6.9 | 2.6 | 3.8 | 4.1 |
| Seaquest | 6.5 | 4.2 | 3.7 | 4.3 |
| YarsRevenge | 6.5 | 6.1 | 3.7 | 4.4 |
| Mean | 6.6 | 3.9 | 3.9 | 4.5 |

Number of filters in convolutional layers one and two: 80/160

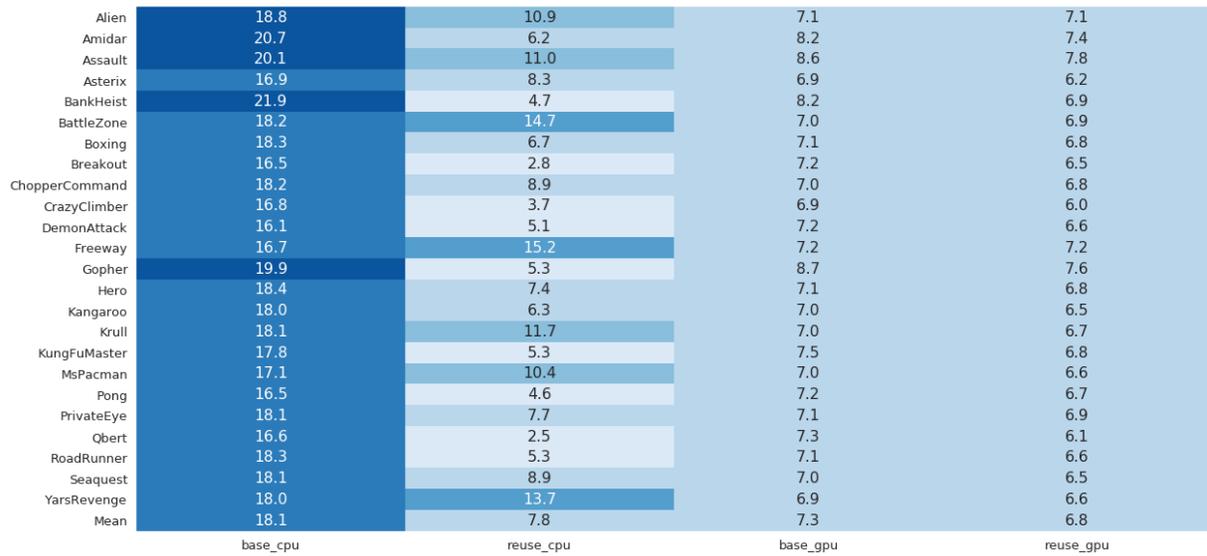

| | base_cpu | reuse_cpu | base_gpu | reuse_gpu |
|---|---|---|---|---|
| Alien | 18.8 | 10.9 | 7.1 | 7.1 |
| Amidar | 20.7 | 6.2 | 8.2 | 7.4 |
| Assault | 20.1 | 11.0 | 8.6 | 7.8 |
| Asterix | 16.9 | 8.3 | 6.9 | 6.2 |
| BankHeist | 21.9 | 4.7 | 8.2 | 6.9 |
| BattleZone | 18.2 | 14.7 | 7.0 | 6.9 |
| Boxing | 18.3 | 6.7 | 7.1 | 6.8 |
| Breakout | 16.5 | 2.8 | 7.2 | 6.5 |
| ChopperCommand | 18.2 | 8.9 | 7.0 | 6.8 |
| CrazyClimber | 16.8 | 3.7 | 6.9 | 6.0 |
| DemonAttack | 16.1 | 5.1 | 7.2 | 6.6 |
| Freeway | 16.7 | 15.2 | 7.2 | 7.2 |
| Gopher | 19.9 | 5.3 | 8.7 | 7.6 |
| Hero | 18.4 | 7.4 | 7.1 | 6.8 |
| Kangaroo | 18.0 | 6.3 | 7.0 | 6.5 |
| Krull | 18.1 | 11.7 | 7.0 | 6.7 |
| KungFuMaster | 17.8 | 5.3 | 7.5 | 6.8 |
| MsPacman | 17.1 | 10.4 | 7.0 | 6.6 |
| Pong | 16.5 | 4.6 | 7.2 | 6.7 |
| PrivateEye | 18.1 | 7.7 | 7.1 | 6.9 |
| Qbert | 16.6 | 2.5 | 7.3 | 6.1 |
| RoadRunner | 18.3 | 5.3 | 7.1 | 6.6 |
| Seaquest | 18.1 | 8.9 | 7.0 | 6.5 |
| YarsRevenge | 18.0 | 13.7 | 6.9 | 6.6 |
| Mean | 18.1 | 7.8 | 7.3 | 6.8 |

Inference: time to run 3000 steps, grayscale, downsampled by a factor of 2 in each dimension

Number of filters in convolutional layers one and two: 20/40

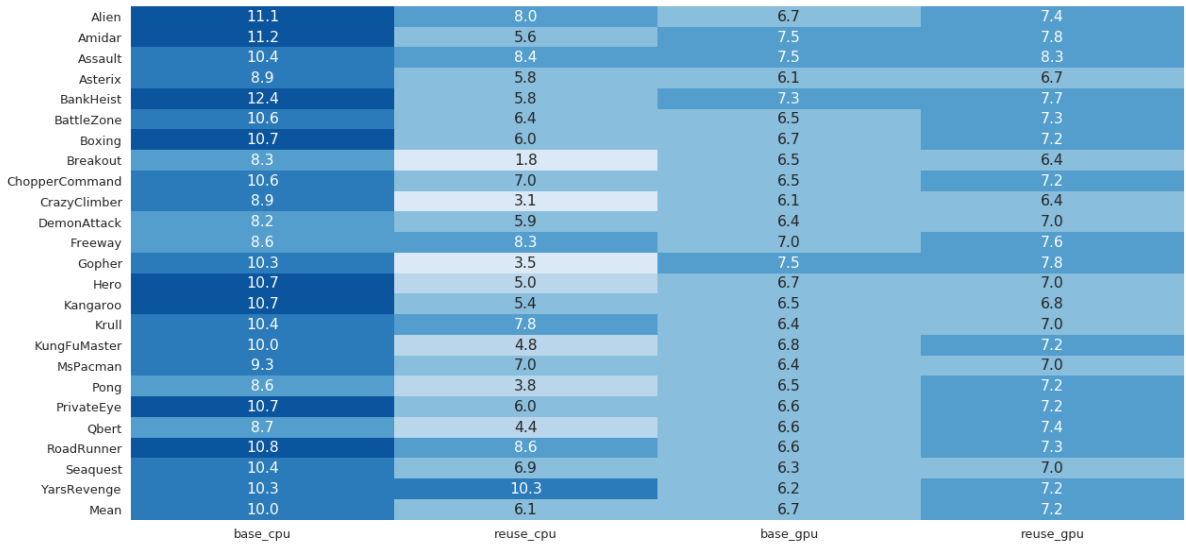

Number of filters in convolutional layers one and two: 40/80

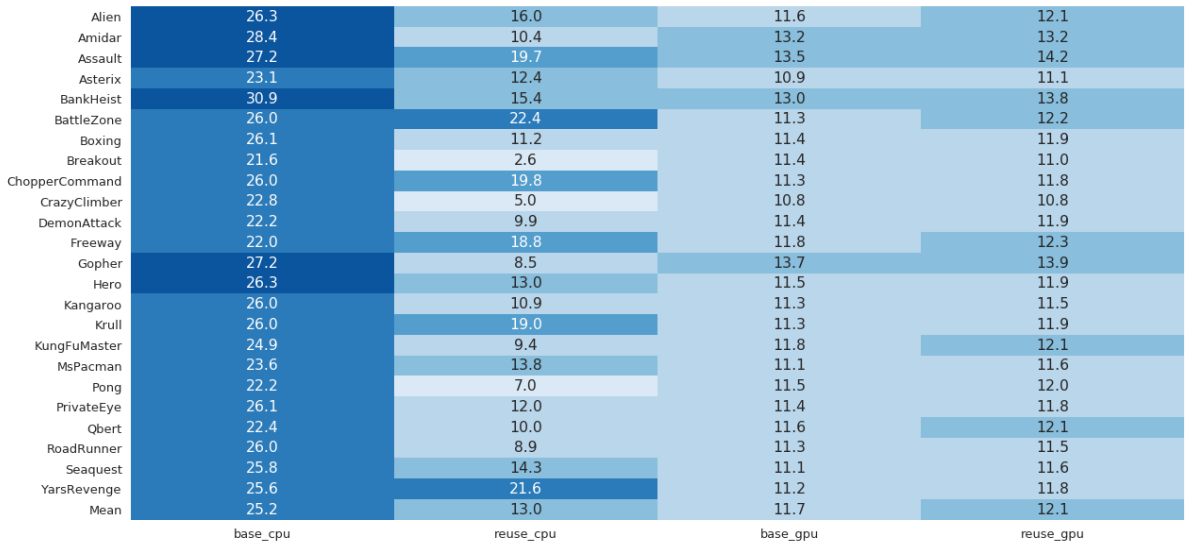

Number of filters in convolutional layers one and two: 80/160

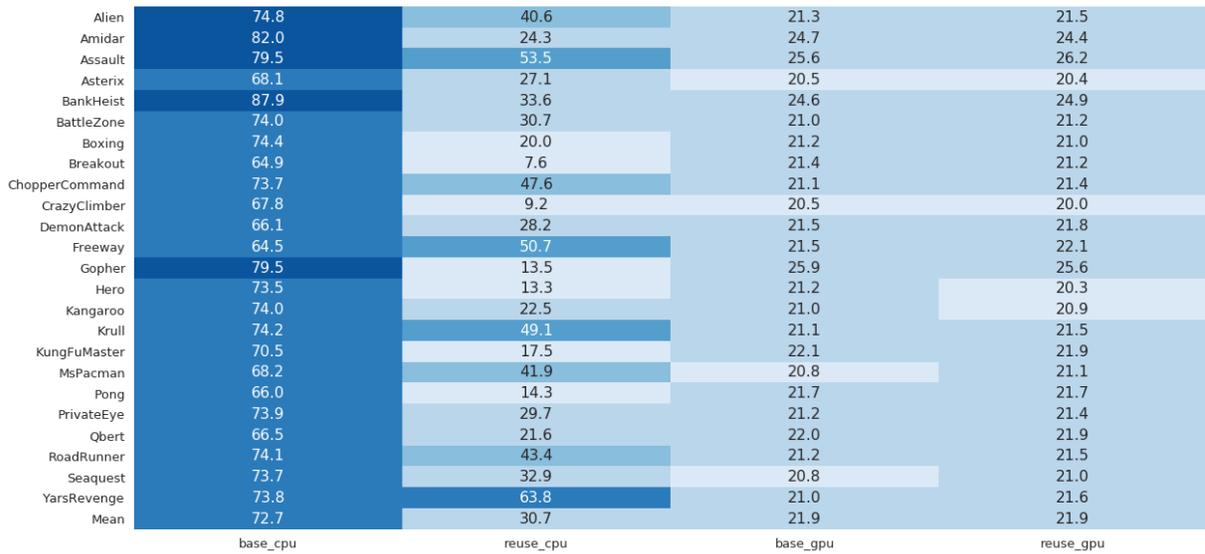

Training: time to run 3000 steps, grayscale, downsampled by a factor of 4 in each dimension

Number of filters in convolutional layers one and two: 20/40

| | base_cpu | reuse_cpu | base_gpu | reuse_gpu |
|---|---|---|---|---|
| Alien | 10.6 | 8.9 | 4.6 | 5.2 |
| Amidar | 9.6 | 6.0 | 4.6 | 4.4 |
| Assault | 7.0 | 5.7 | 3.5 | 4.2 |
| Asterix | 7.5 | 5.9 | 3.9 | 4.3 |
| BankHeist | 7.5 | 5.5 | 3.9 | 4.2 |
| BattleZone | 8.3 | 7.7 | 3.7 | 4.2 |
| Boxing | 9.5 | 7.8 | 4.4 | 4.9 |
| Breakout | 4.0 | 1.7 | 3.1 | 2.7 |
| ChopperCommand | 9.4 | 9.1 | 4.3 | 4.9 |
| CrazyClimber | 3.6 | 2.2 | 2.9 | 2.9 |
| DemonAttack | 3.6 | 3.4 | 2.6 | 3.4 |
| Freeway | 6.7 | 7.3 | 4.3 | 5.3 |
| Gopher | 6.8 | 3.7 | 4.3 | 4.5 |
| Hero | 5.3 | 3.2 | 3.5 | 3.6 |
| Kangaroo | 9.8 | 6.8 | 4.3 | 4.1 |
| Krull | 6.1 | 7.2 | 3.9 | 4.4 |
| KungFuMaster | 9.1 | 6.3 | 4.5 | 4.6 |
| MsPacman | 7.6 | 6.7 | 4.2 | 4.9 |
| Pong | 5.7 | 3.7 | 3.6 | 4.3 |
| PrivateEye | 4.0 | 3.1 | 3.0 | 3.7 |
| Qbert | 6.7 | 3.3 | 4.0 | 3.6 |
| RoadRunner | 9.2 | 7.3 | 4.2 | 4.4 |
| Seaquest | 5.3 | 4.0 | 3.3 | 3.9 |
| YarsRevenge | 7.8 | 7.3 | 4.1 | 4.3 |
| Mean | 7.1 | 5.6 | 3.9 | 4.2 |

Number of filters in convolutional layers one and two: 40/80

| | base_cpu | reuse_cpu | base_gpu | reuse_gpu |
|---|---|---|---|---|
| Alien | 22.5 | 17.2 | 6.0 | 6.5 |
| Amidar | 21.6 | 10.7 | 6.4 | 5.9 |
| Assault | 18.5 | 13.9 | 5.5 | 6.2 |
| Asterix | 16.9 | 12.4 | 5.0 | 5.4 |
| BankHeist | 20.2 | 12.1 | 5.7 | 5.8 |
| BattleZone | 17.1 | 14.3 | 5.1 | 5.7 |
| Boxing | 20.6 | 14.1 | 5.8 | 6.0 |
| Breakout | 7.8 | 1.8 | 4.3 | 4.1 |
| ChopperCommand | 22.9 | 17.3 | 5.8 | 5.9 |
| CrazyClimber | 8.6 | 3.2 | 3.7 | 3.9 |
| DemonAttack | 7.6 | 4.2 | 4.0 | 4.6 |
| Freeway | 14.2 | 14.6 | 5.9 | 6.5 |
| Gopher | 16.9 | 7.1 | 5.5 | 5.2 |
| Hero | 10.2 | 5.0 | 4.4 | 4.5 |
| Kangaroo | 22.5 | 13.3 | 5.8 | 5.2 |
| Krull | 17.6 | 14.2 | 5.0 | 5.6 |
| KungFuMaster | 20.0 | 11.9 | 5.9 | 5.7 |
| MsPacman | 18.1 | 13.8 | 5.5 | 5.9 |
| Pong | 13.2 | 6.3 | 5.0 | 5.5 |
| PrivateEye | 8.0 | 4.2 | 4.1 | 4.8 |
| Qbert | 16.1 | 5.3 | 5.3 | 4.7 |
| RoadRunner | 23.6 | 16.4 | 6.0 | 5.6 |
| Seaquest | 13.2 | 10.9 | 4.8 | 5.1 |
| YarsRevenge | 18.9 | 20.1 | 5.1 | 5.4 |
| Mean | 16.5 | 11.0 | 5.2 | 5.4 |

Number of filters in convolutional layers one and two: 80/160

| | base_cpu | reuse_cpu | base_gpu | reuse_gpu |
|---|---|---|---|---|
| Alien | 52.9 | 37.2 | 9.2 | 9.4 |
| Amidar | 53.9 | 23.1 | 9.7 | 8.7 |
| Assault | 39.2 | 24.8 | | 9.6 |
| Asterix | 43.7 | 26.8 | | 7.9 |
| BankHeist | 33.9 | 14.9 | | 8.6 |
| BattleZone | 48.7 | 36.9 | | |
| Boxing | 50.2 | 26.9 | | 8.6 |
| Breakout | 27.4 | 5.8 | | 6.4 |
| ChopperCommand | 52.5 | 42.1 | | 8.5 |
| CrazyClimber | 18.3 | 5.3 | | 6.2 |
| DemonAttack | 20.3 | 9.7 | | 7.2 |
| Freeway | 38.9 | 36.3 | | |
| Gopher | 37.3 | 11.8 | | 8.5 |
| Hero | 20.3 | 5.8 | | 6.8 |
| Kangaroo | 54.7 | 26.9 | 8.8 | 7.8 |
| Krull | 33.0 | 31.8 | | 8.1 |
| KungFuMaster | 48.4 | 21.9 | | 8.5 |
| MsPacman | 46.8 | 31.7 | 8.3 | 8.5 |
| Pong | 35.4 | 13.4 | 7.9 | 7.9 |
| PrivateEye | 20.0 | 7.7 | | 7.0 |
| Qbert | 40.0 | 9.7 | 8.2 | 7.1 |
| RoadRunner | 52.9 | 26.1 | | 8.3 |
| Seaquest | 28.6 | 17.6 | | 7.6 |
| YarsRevenge | 44.3 | 34.4 | | 8.4 |
| Mean | 39.2 | 22.0 | 8.7 | 8.0 |

Training: time to run 3000 steps, grayscale, downsampled by a factor of 2 in each dimension

Number of filters in convolutional layers one and two: 20/40

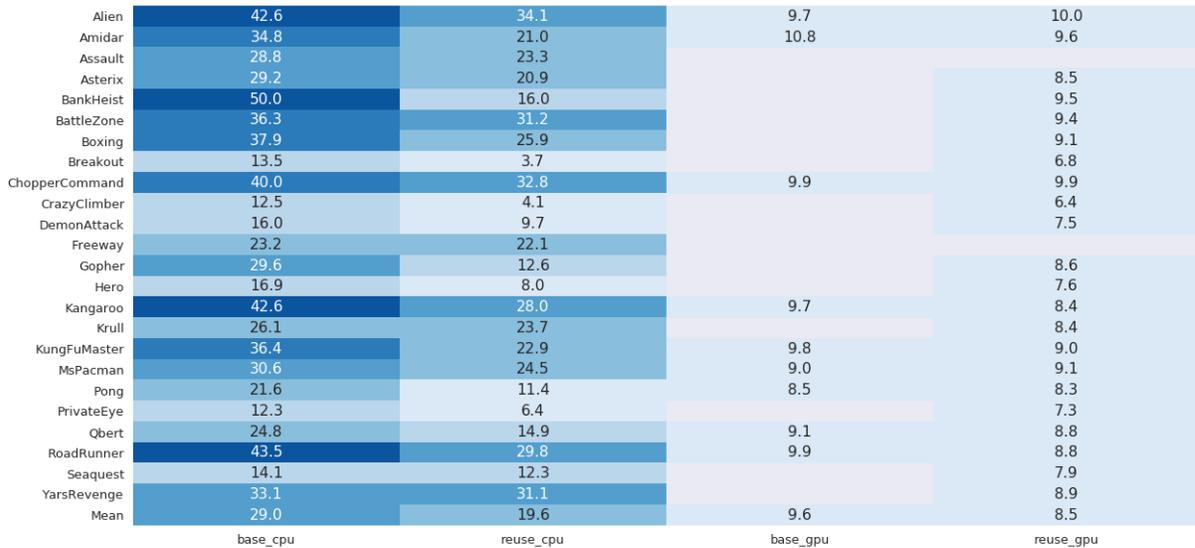

Number of filters in convolutional layers one and two: 40/80

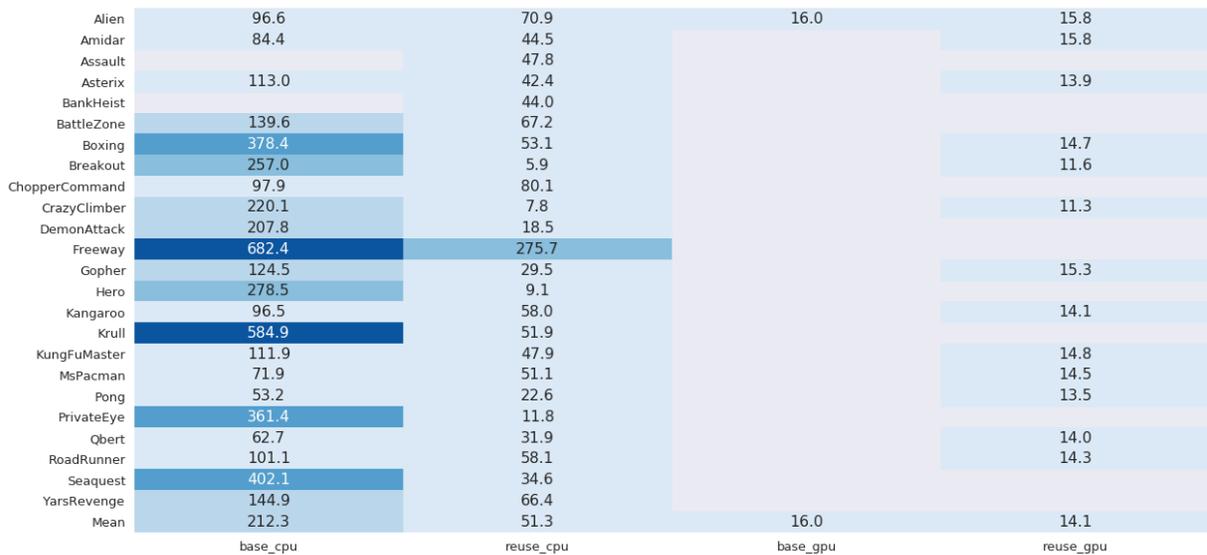